\def\year{2018}\relax
\documentclass[letterpaper]{article} %DO NOT CHANGE THIS
\usepackage{aaai18}  %Required
\usepackage{times}  %Required
\usepackage{helvet}  %Required
\usepackage{courier}  %Required
\usepackage{url}  %Required
\usepackage{graphicx}  %Required
\usepackage[colorlinks=true,linkcolor=blue,urlcolor=blue,citecolor=blue]{hyperref}       % hyperlinks
\usepackage{subfigure}
\usepackage[usenames,dvipsnames,svgnames,table]{xcolor}
\usepackage{arydshln}
\usepackage{amsmath, amsthm, amssymb}

\theoremstyle{remark}

\newcommand{\tabincell}[2]{\begin{tabular}{@{}#1@{}}#2\end{tabular}}

\newcommand{\bs}{\boldsymbol}

\begin{document}

\title{Efficient Architecture Search by Network Transformation}
\author{
    Han Cai$^1$, Tianyao Chen$^1$, Weinan Zhang$^1$\thanks{Correspondence to Weinan Zhang.}, Yong Yu$^1$, Jun Wang$^2$ \\
    $^1$Shanghai Jiao Tong University, $^2$University College London \\
    \{hcai,tychen,wnzhang,yyu\}@apex.sjtu.edu.cn, j.wang@cs.ucl.ac.uk
}
\maketitle
\begin{abstract}
%Deep neural networks have shown effectiveness in many challenging tasks and proved their strong capability in automatically learning good feature representation from raw input. Nonetheless, designing their architectures still requires much human effort. 
Techniques for automatically designing deep neural network architectures such as reinforcement learning based approaches have recently shown promising results. However, their success is based on vast computational resources (e.g. hundreds of GPUs), making them difficult to be widely used. A noticeable limitation is that they still design and train each network from scratch during the exploration of the architecture space, which is highly inefficient. In this paper, we propose a new framework toward efficient architecture search by exploring the architecture space based on the current network and reusing its weights. We employ a reinforcement learning agent as the meta-controller, whose action is to grow the network depth or layer width with function-preserving transformations. As such, the previously validated networks can be reused for further exploration, thus saves a large amount of computational cost. We apply our method to explore the architecture space of the plain convolutional neural networks (no skip-connections, branching etc.) on image benchmark datasets (CIFAR-10, SVHN) with restricted computational resources (5 GPUs). Our method can design highly competitive networks that outperform existing networks using the same design scheme. On CIFAR-10, our model without skip-connections achieves 4.23\% test error rate, exceeding a vast majority of modern architectures and approaching DenseNet. Furthermore, by applying our method to explore the DenseNet architecture space, we are able to achieve more accurate networks with fewer parameters. 
%A key innovation is the employment and extension of a class of function-preserving transformations that allow to initialize the new network to represent the same function as the given network but use different parameterization to be further trained to improve the performance.

% \jun{try to replace "very" and "extremely" etc. with other more former words.}

% However, these methods still train each network from scratch during exploring the architecture space, which results in extremely high computational cost. In this paper, we propose a novel reinforcement learning framework for automatic architecture designing, where the action is to grow the network depth or layer width based on the current network architecture with function preserved. As such, the previously validated networks can be reused for further exploration, thus s a large amount of computational cost. 
\end{abstract}

\section{Introduction}
% several words about the success of NN and then leads to neural achitecture design (related works).
% efficiency problem of current methods
The great success of deep neural networks in various challenging applications \cite{krizhevsky2012imagenet,bahdanau2014neural,silver2016mastering}
has led to a paradigm shift from feature designing to architecture designing, which still remains a laborious task and requires human expertise. In recent years, many techniques for automating the architecture design process have been proposed \cite{snoek2012practical,bergstra2012random,baker2016designing,zoph2016neural,real2017large,negrinho2017deeparchitect}, and promising results of designing competitive models against human-designed models are reported on some benchmark datasets \cite{zoph2016neural,real2017large}. 
Despite the promising results as reported, their success is based on vast computational resources (e.g. hundreds of GPUs), making them difficult to be used in practice for individual researchers, small sized companies, or university research teams. Another key drawback is that they still design and train each network from scratch during exploring the architecture space without any leverage of previously explored networks, which results in high computational resources waste. 

In fact, during the architecture design process, many slightly different networks are trained for the same task. Apart from their final validation performances that are used to guide exploration, we should also have access to their architectures, weights, training curves etc., which contain abundant knowledge and can be leveraged to accelerate the architecture design process just like human experts \cite{chen2015net2net,klein2016learning}. Furthermore, there are typically many well-designed architectures, by human or automatic architecture designing methods, that have achieved good performances at the target task. Under restricted computational resources limits, instead of totally neglecting these existing networks and exploring the architecture space from scratch (which does not guarantee to result in better performance architectures), a more economical and efficient alternative could be exploring the architecture space based on these successful networks and reusing their weights. 

In this paper, we propose a new framework, called EAS, Efficient Architecture Search, where the meta-controller explores the architecture space by \emph{network transformation} operations such as widening a certain layer (more units or filters), inserting a layer, adding skip-connections etc., given an existing network trained on the same task. To reuse weights, we consider the class of function-preserving transformations \cite{chen2015net2net} that allow to initialize the new network to represent the same function as the given network but use different parameterization to be further trained to improve the performance, which can significantly accelerate the training of the new network especially for large networks. Furthermore, we combine our framework with recent advances of reinforcement learning (RL) based automatic architecture designing methods \cite{baker2016designing,zoph2016neural}, and employ a RL based agent as the meta-controller. 

Our experiments of exploring the architecture space of the plain convolutional neural networks (CNNs), which purely consists of convolutional, fully-connected and pooling layers without skip-connections, branching etc., on image benchmark datasets (CIFAR-10, SVHN), show that EAS with limited computational resources (5 GPUs) can design competitive architectures. The best plain model designed by EAS on CIFAR-10 with standard data augmentation achieves 4.23\% test error rate, even better than many modern architectures that use skip-connections. We further apply our method to explore the DenseNet \cite{huang2016densely} architecture space, and achieve 4.66\% test error rate on CIFAR-10 without data augmentation and 3.44\% on CIFAR-10 with standard data augmentation, surpassing the best results given by the original DenseNet while still maintaining fewer parameters. 

%\jun{suggest to have some explicity wording like "Our key contributions lie in.." for novelty claims.}
% The experiments on CIFAR-10 show that our agent with very limited computational resources compared to existing automatic architecture designing methods, is able to design very competitive models against both human-designed models and automatically designed models. 

% experiments and etc.

\section{Related Work and Background}\label{sec:relate} %\jun{no need sub-section titles?}
% automatic architecture design: Bayesian optimization, neuro-evolution, reinforcement learning 
\subsubsection{Automatic Architecture Designing}\label{para:aad}
There is a long standing study on automatic architecture designing. 
Neuro-evolution algorithms which mimic the evolution processes in the nature, are one of the earliest automatic architecture designing methods \cite{miller1989designing,stanley2002evolving}. 
%, yet are usually unable to match the performance of human-designed models \cite{verbancsics2013generative}
Authors in \cite{real2017large} used neuro-evolution algorithms to explore a large CNN architecture space and achieved networks which can match performances of human-designed models. 
%showed that neuro-evolution algorithms are capable of constructing large networks which can match the performances of human-designed models.
%However, they are actually search-based methods, thus require enormous computational power to work well, which makes them less practical at a large scale. 
In parallel, automatic architecture designing has also been studied in the context of Bayesian optimization \cite{bergstra2012random,domhan2015speeding,mendoza2016towards}. 
%but these methods only search models from a fixed-length space and are unable to generate variable-length network architectures.
Recently, reinforcement learning is introduced in automatic architecture designing and has shown strong empirical results. Authors in
\cite{baker2016designing} presented a Q-learning agent to sequentially pick CNN layers; authors in \cite{zoph2016neural} used an auto-regressive recurrent network to generate a variable-length string that specifies the architecture of a neural network and trained the recurrent network with policy gradient.

As the above solutions rely on designing or training networks from scratch, significant computational resources have been wasted during the construction. In this paper, we aim to address the efficiency problem. Technically, we allow to reuse the existing networks trained on the same task and take network transformation actions. Both function-preserving transformations and an alternative RL based meta-controller are used to explore the architecture space. Moreover, we notice that there are some complementary techniques, such as learning curve prediction \cite{klein2016learning}, for improving the efficiency, which can be combined with our method. 

% network transformation: net2net, network compression
\subsubsection{Network Transformation and Knowledge Transfer}\label{para:net_trans}
Generally, any modification to a given network can be viewed as a network transformation operation. In this paper, since our aim is to utilize knowledge stored in previously trained networks, we focus on identifying the kind of network transformation operations that would be able to reuse pre-existing models. The idea of reusing pre-existing models or knowledge transfer between neural networks has been studied before. Net2Net technique introduced in \cite{chen2015net2net} describes two specific function-preserving transformations, namely Net2WiderNet and Net2DeeperNet, which respectively initialize a wider or deeper student network to represent the same functionality of the given teacher network and have proved to significantly accelerate the training of the student network especially for large networks. 
Similar function-preserving schemes have also been proposed in ResNet particularly for training very deep architectures \cite{he2016deep}.
Additionally, the network compression technique presented in \cite{han2015learning} prunes less important connections (low-weight connections) in order to shrink the size of neural networks without reducing their accuracy. 

In this paper, instead, we focus on utilizing such network transformations to reuse pre-existing models to efficiently and economically explore the architecture space for automatic architecture designing. 
%We mainly consider Net2Net operations while other kind of network transformation operations can be easily incorporated under the same framework. \jun{why we limit our self? any benefit of N2N???if the novelty is not about using N2N then what it is?a bit risky about novelty here from the last sentence.}
% To the best of our knowledge, it is the first work trying to learn an intelligent agent to perform network transformation operations for automatic architecture designing.

% reinforcement learning and meta-learning
\subsubsection{Reinforcement Learning Background}\label{para:rl-meta}
%Our work is related to meta-learning or learning to learn \cite{vilalta2002perspective}, the idea of using acquired knowledge from one task to improve learning in a future task.
%\tianyao{This description is for transfer learning, I think..?}. \han{I slightly modify the description. There are many views of meta-learning. Here I use a similar description in \cite{zoph2016neural}.}
%More closely related, \cite{andrychowicz2016learning} used a long short-term memory (LSTM) network to learn to optimize another network and \cite{wang2016learning} used RL techniques to train a recurrent neural network to find update policies for another network.
Our meta-controller in this work is based on RL \cite{sutton1998reinforcement}, techniques for training the agent to maximize the cumulative reward when interacting with an environment \cite{cai2017real}. We use the REINFORCE algorithm \cite{williams1992simple} similar to \cite{zoph2016neural} for updating the meta-controller, while other advanced policy gradient methods \cite{kakade2002natural,schulman2015trust} can be applied analogously. Our action space is, however, different with that of \cite{zoph2016neural} or any other RL based approach \cite{baker2016designing}, as our actions are the network transformation operations like adding, deleting, widening, etc., while others are specific configurations of a newly created network layer on the top of preceding layers. 
%As such, we significantly reduce the exploration space. 
Specifically, we model the automatic architecture design procedure as a sequential decision making process, where the state is the current network architecture and the action is the corresponding network transformation operation. After $T$ steps of network transformations, the final network architecture, along with its weights transferred from the initial input network, is then trained in the real data to get the validation performance to calculate the reward signal, which is further used to update the meta-controller via policy gradient algorithms to maximize the expected validation performances of the designed networks by the meta-controller.

%Compared to the two previous RL based automatic architecture designing methods \cite{baker2016designing,zoph2016neural} that pose schemes with sequentially picking the next layer, which restrict the meta-controller to start from scratch at the beginning of each design process without reusing any weights of pre-existing network and add a layer at the top given existing layers below at each step, our proposed method EAS explores a totally distinct architecture design scheme where
%can explore the architecture space much more flexibly. 

%\jun{I would expect you spend more texts on comparing your approach to other RL based network architecture search???}
%Our work is based on RL, techniques for training the agent to maximize the cumulative reward in a sequential agent-environment interaction process. Similar to \cite{zoph2016neural}, we use a neural network to represent the policy of the agent and train the network with policy gradient methods \cite{williams1992simple,kakade2002natural,schulman2015trust}. 

\section{Architecture Search by Net Transformation}\label{sec:method}
% formulation and models

In this section, we first introduce the overall framework of our meta-controller, and then show how each specific network transformation decision is made under it. We later extend the function-preserving transformations to the DenseNet \cite{huang2016densely} architecture space where directly applying the original Net2Net operations can be problematic since the output of a layer will be fed to all subsequent layers. 

%In this section, we first introduce a general framework for learning an intelligent agent to automatically take network transformation actions. We will further show how the agent can be trained via policy gradient methods to maximize the expected validation performances of the result networks. Finally, we will discuss the connections of our work to previous automatic architecture designing approaches. 
% In this section, we firstly describe a general framework for learning an intelligent agent to automatically perform network transformation operations, which consists of a recurrent network for encoding the given network architecture (learning the state representation) and separate actor components for taking each type of network transformation operation. 
% We further describe two specific actor components, Net2Wider actor and Net2Deeper actor which correspond to the two specific Net2Net operations, Net2WiderNet and Net2DeeperNet respectively. We will show how the agent can be trained via policy gradient methods to maximize the expected validation performances of final result architectures. Finally, we discuss connections of our work to previous works. 

%\subsection{Learning to Perform Network Transformation}
% Encoder-Decode Framework

\begin{figure}[t]
	\centering
	\includegraphics[width=\columnwidth]{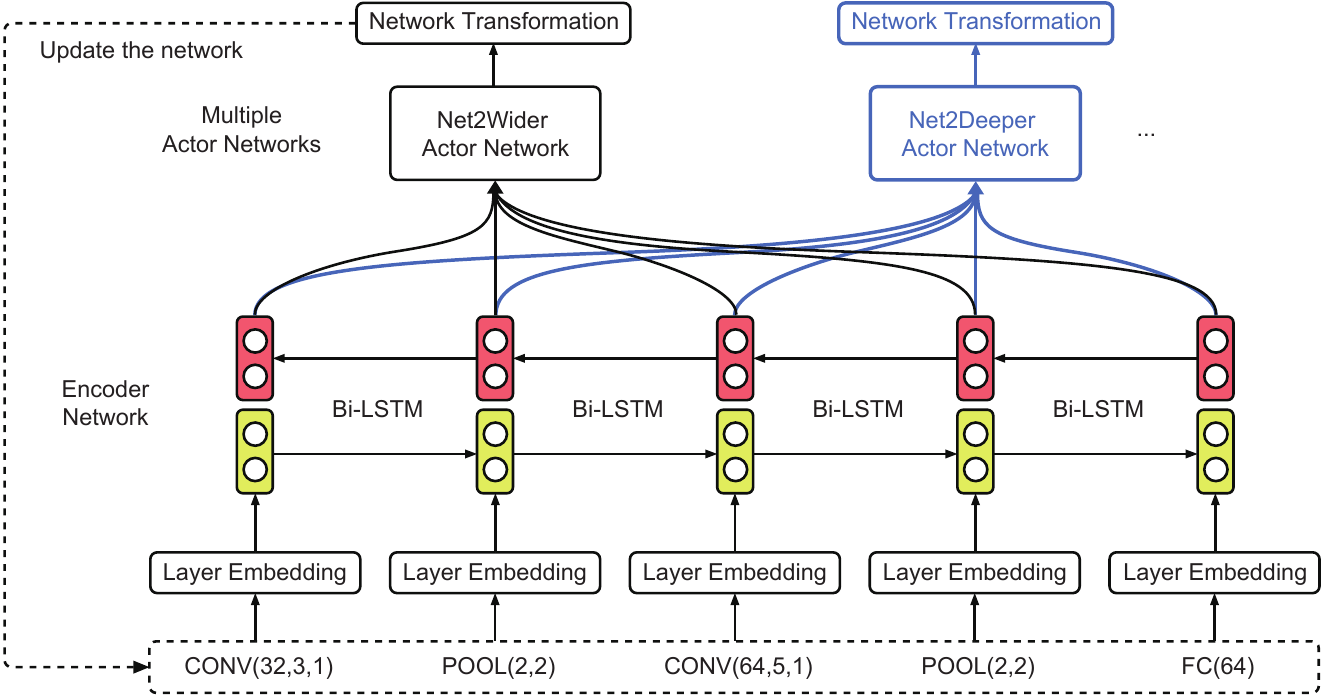}
	\caption{Overview of the RL based meta-controller in EAS, which consists of an encoder network for encoding the architecture and multiple separate actor networks for taking network transformation actions.}
	\label{fig:rl-meta-frame}
\end{figure}

% Encoder
We consider learning a meta-controller to generate network transformation actions given the current network architecture, which is specified with a variable-length string \cite{zoph2016neural}. To be able to generate various types of network transformation actions while keeping the meta-controller simple, we use an encoder network to learn a low-dimensional representation of the given architecture, which is then fed into each separate actor network to generate a certain type of network transformation actions. Furthermore, to handle variable-length network architectures as input and take the whole input architecture into consideration when making decisions, the encoder network is implemented with a bidirectional recurrent network \cite{schuster1997bidirectional} with an input embedding layer. The overall framework is illustrated in Figure~\ref{fig:rl-meta-frame}, which is an analogue of end-to-end sequence to sequence learning \cite{sutskever2014sequence,bahdanau2014neural}. 

\subsection{Actor Networks}
% Decoder
Given the low dimensional representation of the input architecture, each actor network makes necessary decisions for taking a certain type of network transformation actions. In this work, we introduce two specific actor networks, namely Net2Wider actor and Net2Deeper actor which correspond to Net2WiderNet and Net2DeeperNet respectively. 

\subsubsection{Net2Wider Actor}
Net2WiderNet operation allows to replace a layer with a wider layer, meaning more units for fully-connected layers, or more filters for convolutional layers, while preserving the functionality. For example, consider a convolutional layer with kernel $\bs{K}_l$ whose shape is $(k^l_w, k^l_h, f^l_i, f^l_o)$ where $k^l_w$ and $k^l_h$ denote the filter width and height, while $f^l_i$ and $f^l_o$ denote the number of input and output channels. To replace this layer with a wider layer that has $\hat{f}^l_o$ ($> f^l_o$) output channels, we should first introduce a random remapping function $G_l$, which is defined as
{\small
\begin{equation}
\label{eq:remapping}
    G_l(j) = \begin{cases}
        j & \!\! 1 \leq j \leq f^l_o \\
        \text{random sample from}~\{1, \cdots, f^l_o\} & \!\! f^l_o \! < \! j \! \leq \! \hat{f}^l_o
    \end{cases}.
\end{equation}
}With the remapping function $G_l$, we have the new kernel $\hat{\bs{K}}_l$ for the wider layer with shape $(k^l_w, k^l_h, f^l_i, \hat{f}^l_o)$
{\small
\begin{equation}
\label{eq:wider_kernel}
    \hat{\bs{K}}_l [x, y, i, j] = \bs{K}_l [x, y, i, G_l(j)].
\end{equation}
}As such, the first $f^l_o$ entries in the output channel dimension of $\hat{\bs{K}}_l$ are directly copied from $\bs{K}_l$ while the remaining $\hat{f}^l_o - f^l_o$ entries are created by choosing randomly as defined in $G_l$. 
Accordingly, the new output of the wider layer is $\hat{\bs{O}}_l$ with $\hat{\bs{O}}_l(j) = \bs{O}_l(G_l(j))$, where $\bs{O}_l$ is the output of the original layer and we only show the channel dimension to make the notation simpler. 

To preserve the functionality, the kernel $\bs{K}_{l+1}$ of the next layer should also be modified due to the replication in its input. The new kernel $\hat{\bs{K}}_{l+1}$ with shape $(k^{l+1}_w, k^{l+1}_h, \hat{f}^{l+1}_i = \hat{f}^l_o, f^{l+1}_o)$ is given as 
{\small
\begin{equation}
\label{eq:prev_widen}
    \hat{\bs{K}}_{l+1} [x, y, j, k] = \frac{\bs{K}_{l+1} [x, y, G_l(j), k]}{\big| \{z | G_l(z) = G_l(j)\} \big|}.
\end{equation}
}For further details, we refer to the original Net2Net work \cite{chen2015net2net}.

\begin{figure}[t]
	\centering
	\includegraphics[width=0.7\columnwidth]{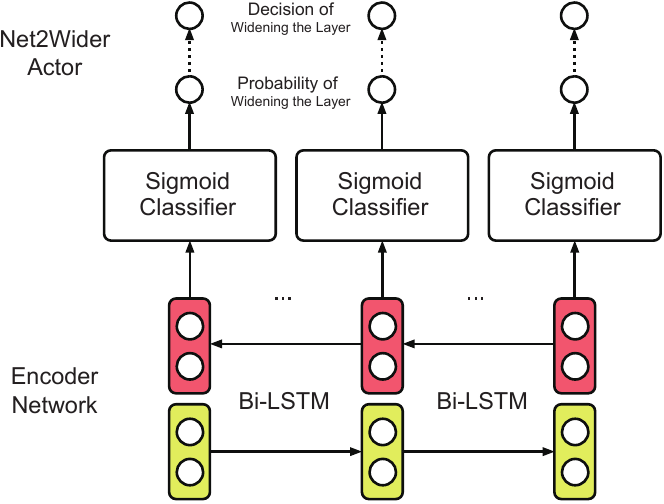}
	\caption{Net2Wider actor, which uses a shared sigmoid classifier to simultaneously determine whether to widen each layer based on its hidden state given by the encoder network.}
    \label{fig:wider-nets}
\end{figure}

In our work, to be flexible and efficient, the Net2Wider actor simultaneously determines whether each layer should be extended. Specifically, for each layer, this decision is carried out by a shared sigmoid classifier given the hidden state of the layer learned by the bidirectional encoder network.
%\tianyao{Traditionally, Net2WiderNet operation means to make a specific layer wider. In our work, on the other hand, our agent's Net2Wider action simultaneously determines whether each layer should be extended. This action gives a binary list to indicate this action. This binary list is acquired by a sigmoid classifier given the hidden state of the layer learned by the bidirectional encoder network. We choose this method, instead of widen one layer each time, since the this method is more efficient, which can widen several layers at one action, and it includes the traditional case when it decides to widen only one layer.}\han{Correspondingly, Net2Wider actor makes binary decision for each convolutional or fully-connected layer on whether to widen the layer, which is carried out by a sigmoid classifier given the hidden state of the layer learned by the bidirectional encoder network.}
Moreover, we follow previous work and search the number of filters for convolutional layers and units for fully-connected layers in a discrete space. Therefore, if the Net2Wider actor decides to widen a layer, the number of filters or units of the layer increases to the next discrete level, e.g. from 32 to 64. The structure of Net2Wider actor is shown in Figure~\ref{fig:wider-nets}. 

%\subfigure[Net2Deeper Network]{\includegraphics[height=0.3\columnwidth]{figures/rl-meta-deeper-2.pdf}\label{fig:net2deeper}} 

%Furthermore, while Net2WiderNet operation in \cite{chen2015net2net} only modifies the number of filters or units of a layer, we find a simple way to increase the filter size of a convolutional layer with the functionality preserved. Consider the most commonly used setting, i.e. the width and the height of the filter are odd and the zero-padding parameter is set to preserve the spatial size of the input volume, for any kernel $K$ with width $k_x$, height $k_y$, input channel $f_i$ and output channel $f_o$, expressed as a tensor of shape $(k_x, k_y, f_i, f_o)$, it can be widen to another kernel $K'$ of shape $(k_x', k_y', f_i, f_o)$ where $k_x, k_x', k_y$ and $k_y'$ are odd and $k_y' > k_y, k_x' > k_x$, with functionality preserved as long as:
%\begin{equation}\small
%	K'[i, j, p, l] = \begin{cases}
%		K[i-\delta_x, j-\delta_y, p, l] & \delta_x \leq i < k_x'-\delta_x\\ & \delta_y \leq j < k_y'-\delta_y \\
%		0 & \text{otherwise}
%	\end{cases},
%\end{equation}
%where $\delta_x = \frac{k_x'-k_x}{2}$ and $\delta_y = \frac{k_y'-k_y}{2}$. 
%As such, we can modify both the filter number of a convolutional layer and its filter size while preserving the functionality, thereby widen the search space. 
%\weinan{Why such a modification way is better? Please describe its advantages here.} \han{the main advantage is widening the search space.}

\subsubsection{Net2Deeper Actor}
\begin{figure}[t]
	\centering
	\includegraphics[width=0.75\columnwidth]{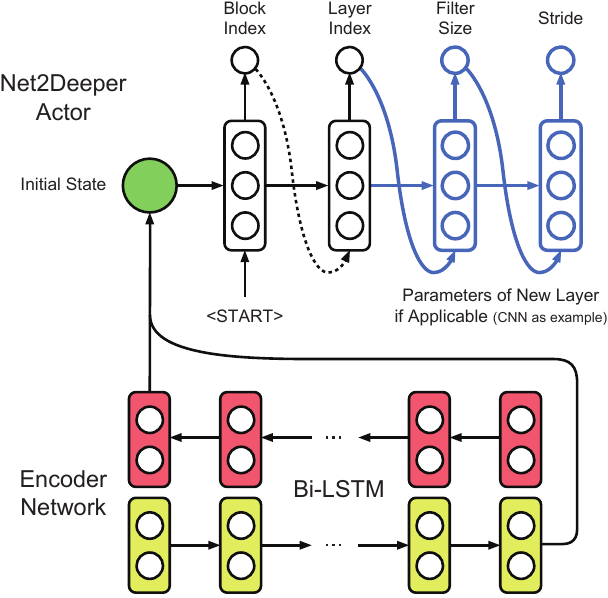}
	\caption{Net2Deeper actor, which uses a recurrent network to sequentially determine where to insert the new layer and corresponding parameters for the new layer based on the final hidden state of the encoder network given the input architecture.}
    \label{fig:deeper-nets}
\end{figure}

Net2DeeperNet operation allows to insert a new layer that is initialized as adding an identity mapping between two layers so as to preserve the functionality. For a new convolutional layer, the kernel is set to be identity filters while for a new fully-connected layer, the weight matrix is set to be identity matrix. Thus the new layer is set with the same number of filters or units as the layer below at first, and could further get wider when Net2WiderNet operation is performed on it. To fully preserve the functionality, Net2DeeperNet operation has a constraint on the activation function $\phi$, i.e. $\phi$ must satisfy $\phi(\bs{I} \phi(\bs{v})) = \phi(\bs{v})$ for all vectors $\bs{v}$.
%, which means the  put on the activation of previous layer should produce the same result.
%\tianyao{The new layer is not constrained to be as wide as the layer below, since we can further perform Net2WiderNet operation on the new layer.}\han{Since we can further perform Net2WiderNet operation on the new layer, so actually the new layer is not constrained to be as wide as the layer below.}
%\tianyao{For Net2Deeper operation, we do not have many choices on activation function.}
%\tianyao{which means the non-linearity put on the activation of previous layer should produce the same result}. 
This property holds for rectified linear activation (ReLU) but fails for sigmoid and tanh activation. However, we can still reuse weights of existing networks  with sigmoid or tanh activation, which could be useful compared to random initialization. Additionally, when using batch normalization \cite{ioffe2015batch}, we need to set output scale and output bias
%\tianyao{the multiplication coefficients and the addition coefficients} \han{output scale and output bias} 
of the batch normalization layer to undo the normalization, rather than initialize them as ones and zeros. Further details about the Net2DeeperNet operation is provided in the original paper \cite{chen2015net2net}.

The structure of the Net2Deeper actor is shown in Figure~\ref{fig:deeper-nets}, which is a recurrent network whose hidden state is initialized with the final hidden state of the encoder network. 
Similar to previous work \cite{baker2016designing}, we allow the Net2Deeper actor to insert one new layer at each step. Specifically, we divide a CNN architecture into several blocks according to the pooling layers and Net2Deeper actor sequentially determines which block to insert the new layer, a specific index within the block and parameters of the new layer. For a new convolutional layer, the agent needs to determine the filter size and the stride while for a new fully-connected layer, no parameter prediction is needed. In CNN architectures, any fully-connected layer should be on the top of all convolutional and pooling layers. To avoid resulting in unreasonable architectures, if the Net2Deeper actor decides to insert a new layer after a fully-connected layer or the final global average pooling layer, the new layer is restricted to be a fully-connected layer, otherwise it must be a convolutional layer. 
% convolutional layer or pooling layer, the new layer is restricted to be a convolutional layer, and if the agent decides to insert a new layer after a fully-connected layer, the new layer must be a fully-connected layer. 

%Specifically the Net2Deeper actor first determines the index of the layer where the new layer is inserted in and then determines parameters of the new layer, which are carried out by separate softmax classifiers. For a new convolutional layer, the agent needs to determine the filter size and the stride while for a new fully-connected layer, no parameter prediction is needed. In CNN architectures, any fully-connected layer should be on the top of all convolutional and pooling layers. To avoid resulting in unreasonable architectures, if the Net2Deeper actor decides to insert a new layer after a convolutional layer or pooling layer, the new layer is restricted to be a convolutional layer, and if the agent decides to insert a new layer after a fully-connected layer, the new layer must be a fully-connected layer. 
% \weinan{So you must have a fc layer at the beginning?} \han{no, we can have a fc layer at the beginning and no fc layer is also ok. But if you want fc layer in your search space, you must have a fc layer at the beginning.}

\subsection{Function-preserving Transformation for DenseNet}
The original Net2Net operations proposed in \cite{chen2015net2net} are discussed under the scenarios where the network is arranged layer-by-layer, i.e. the output of a layer is only fed to its next layer. As such, in some modern CNN architectures where the output of a layer would be fed to multiple subsequent layers, such as DenseNet \cite{huang2016densely}, directly applying the original Net2Net operations can be problematic. In this section, we introduce several extensions to the original Net2Net operations to enable function-preserving transformations for DenseNet. 

Different from the plain CNN, in DenseNet, the $l^{th}$ layer would receive the outputs of all preceding layers as input, which are concatenated on the channel dimension, denoted as $[\bs{O}_0, \bs{O}_1, \cdots, \bs{O}_{l-1}]$, while its output $\bs{O}_l$ would be fed to all subsequent layers. 

Denote the kernel of the $l^{th}$ layer as $\bs{K}_l$ with shape $(k^l_w, k^l_h, f^l_i, f^l_o)$. To replace the $l^{th}$ layer with a wider layer that has $\hat{f}^l_o$ output channels while preserving the functionality, the creation of the new kernel $\hat{\bs{K}}_l$ in the $l^{th}$ layer is the same as the original Net2WiderNet operation (see Eq.~(\ref{eq:remapping}) and Eq.~(\ref{eq:wider_kernel})). As such, the new output of the wider layer is $\hat{\bs{O}}_l$ with $\hat{\bs{O}}_l(j) = \bs{O}_l(G_l(j))$, where $G_l$ is the random remapping function as defined in Eq.~(\ref{eq:remapping}). 
Since the output of the $l^{th}$ layer will be fed to all subsequent layers in DenseNet, the replication in $\hat{\bs{O}}_l$ will result in replication in the inputs of all layers after the $l^{th}$ layer. As such, instead of only modifying the kernel of the next layer as done in the original Net2WiderNet operation, we need to modify the kernels of all subsequent layers in DenseNet. For the $m^{th}$ layer where $m > l$, its input becomes $[\bs{O}_0, \cdots, \bs{O}_{l-1}, \hat{\bs{O}}_l, \bs{O}_{l+1}, \cdots, \bs{O}_{m-1}]$ after widening the $l^{th}$ layer, thus from the perspective of $m^{th}$ layer, the equivalent random remapping function $\hat{G}_m$ can be written as
{\small
\begin{equation}
\label{eq:eqv_remapping}
	\hat{G}_m(j) \! = \!
    \begin{cases}
		j & 1 \leq j \leq f^{0:l}_o \\
		f^{0:l}_o \! + \! G_l(j) & f^{0:l}_o \! < \! j \! \leq \! f^{0:l}_o \! + \! \hat{f}^l_o  \\
		j - \hat{f}^l_o \! + \! f^l_o & f^{0:l}_o + \hat{f}^l_o < j \leq f^{0:m}_o \! + \! \hat{f}^l_o \! - \! f^l_o \\ 
	\end{cases},
\end{equation}
}where $f^{0:l}_o = \sum_{v=0}^{l-1} f^v_o$ is the number of input channels for the $l^{th}$ layer, the first part corresponds to $[\bs{O}_0, \cdots, \bs{O}_{l-1}]$, the second part corresponds to $[\hat{\bs{O}}_l]$, and the last part corresponds to $[\bs{O}_{l+1}, \cdots, \bs{O}_{m-1}]$. 
A simple example of $\hat{G}_m$ is given as 
{\small
\begin{align*}
    \hat{G}_m &: \{1, \cdots, 5, \overbrace{6, 7, 8, 9}^{\hat{\bs{O}}_l}, 10, 11\} \rightarrow \{1, \cdots, 5, \overbrace{6, 7, 6, 6}^{\hat{\bs{O}}_l}, 8, 9\} \\
              & \quad \text{where } G_l: \{1, 2, 3, 4\} \rightarrow \{1, 2, 1, 1\}.
\end{align*}
}Accordingly the new kernel of $m^{th}$ layer can be given by Eq.~(\ref{eq:prev_widen}) with $G_l$ replaced with $\hat{G}_m$.

To insert a new layer in DenseNet, suppose the new layer is inserted after the $l^{th}$ layer. Denote the output of the new layer as $\bs{O}_{\text{new}}$, and its input is $[\bs{O}_0, \bs{O}_1, \cdots, \bs{O}_{l}]$. Therefore, for the $m^{th}$ $(m > l)$ layer, its new input after the insertion is $[\bs{O}_0, \bs{O}_1, \cdots, \bs{O}_{l}, \bs{O}_{\text{new}}, \bs{O}_{l+1}, \cdots, \bs{O}_{m-1}]$. To preserve the functionality, similar to the Net2WiderNet case, $\bs{O}_{\text{new}}$ should be the replication of some entries in $[\bs{O}_0, \bs{O}_1, \cdots, \bs{O}_{l}]$. It is possible, since the input of the new layer is $[\bs{O}_0, \bs{O}_1, \cdots, \bs{O}_{l}]$. 
Each filter in the new layer can be represented with a tensor, denoted as $\hat{\bs{F}}$ with shape $(k^{\text{new}}_w, k^{\text{new}}_h, f^{\text{new}}_i = f^{0:l+1}_o) $, where $k^{\text{new}}_w$  and $k^{\text{new}}_h$ denote the width and height of the filter, and $f^{\text{new}}_i$ is the number of input channels. To make the output of $\hat{\bs{F}}$ to be a replication of the $n^{th}$ entry in $[\bs{O}_0, \bs{O}_1, \cdots, \bs{O}_{l}]$,
we can set $\hat{\bs{F}}$ (using the special case that $k^{\text{new}}_w$ = $k^{\text{new}}_h$ = 3 for illustration) as the following
{\small
\begin{equation}
    \hat{\bs{F}} [x, y, n] = 
            \begin{bmatrix}
            0 & 0 & 0 \\
            0 & 1 & 0 \\
            0 & 0 & 0 \\
            \end{bmatrix},
\end{equation}
}while all other values in $\hat{\bs{F}}$ are set to be 0.
Note that $n$ can be chosen randomly from $\{1, \cdots, f^{0:l+1}_o\}$ for each filter. 
After all filters in the new layer are set, we can form an equivalent random remapping function for all subsequent layers as is done in Eq.~(\ref{eq:eqv_remapping}) and modify their kernels accordingly.

\begin{table*}[t]
	\centering
	\caption{Simple start point network. C$(n, f, l)$ denotes a convolutional layer with $n$ filters, filter size $f$ and stride $l$; P$(f, l, \text{MAX})$ and P$(f, l, \text{AVG})$ denote a max and an average pooling layer with filter size $f$ and stride $l$ respectively; FC$(n)$ denotes a fully-connected layer with $n$ units; SM$(n)$ denotes a softmax layer with $n$ output units.}
    \label{tab:start_net}
    \resizebox{0.75\textwidth}{!}{  
	\begin{tabular}{| l | c |}
        \hline
		Model Architecture & Validation Accuracy (\%)\\
        \hline
        \tabincell{l}{C(16, 3, 1), P(2, 2, MAX), C(32, 3, 1), P(2, 2, MAX), C(64, 3, 1), \\ P(2, 2, MAX), C(128, 3, 1), P(4, 4, AVG), FC(256), SM(10)} & 87.07 \\
		\hline
	\end{tabular}
    }
\end{table*}

\section{Experiments and Results}
% make the intentions of the experiemnts clear, emphasize that we publicize code and models
In line with the previous work \cite{baker2016designing,zoph2016neural,real2017large}, we apply the proposed EAS on image benchmark datasets (CIFAR-10 and SVHN) to explore high performance CNN architectures for the image classification task\footnote{Experiment code and discovered top architectures along with weights: \url{https://github.com/han-cai/EAS}}. Notice that the performances of the final designed models largely depend on the architecture space and the computational resources. In our experiments, we evaluate EAS in two different settings. In all cases, we use restricted computational resources (5 GPUs) compared to the previous work such as \cite{zoph2016neural} that used 800 GPUs. In the first setting, we apply EAS to explore the plain CNN architecture space, which purely consists of convolutional,  pooling  and fully-connected layers. While in the second setting, we apply EAS to explore the DenseNet architecture space. 

\subsection{Image Datasets}
\subsubsection{CIFAR-10} The CIFAR-10 dataset \cite{krizhevsky2009learning} consists of 50,000 training images and 10,000 test images.
We use a standard data augmentation scheme that is widely used for CIFAR-10 \cite{huang2016densely}, and denote the augmented dataset as C10+ while the original dataset is denoted as C10. For preprocessing, we normalized the images using the channel means and standard deviations. Following the previous work \cite{baker2016designing,zoph2016neural}, we randomly sample 5,000 images from the training set to form a validation set while using the remaining 45,000 images for training during exploring the architecture space. 

\subsubsection{SVHN} The Street View House Numbers (SVHN) dataset \cite{netzer2011reading} contains 73,257 images in the original training set, 26,032 images in the test set, and 531,131 additional images in the extra training set. For preprocessing, we divide the pixel values by 255 and do not perform any data augmentation, as is done in \cite{huang2016densely}. 
We follow \cite{baker2016designing} and use the original training set during the architecture search phase with 5,000 randomly sampled images as the validation set, while training the final discovered architectures using all the training data, including the original training set and extra training set. 

\subsection{Training Details}
For the meta-controller, we use a one-layer bidirectional LSTM with 50 hidden units as the encoder network (Figure \ref{fig:rl-meta-frame}) with an embedding size of 16, and train it with the ADAM optimizer \cite{kingma2014adam}. 

At each step, the meta-controller samples 10 networks by taking network transformation actions. Since the sampled networks are not trained from scratch but we reuse weights of the given network in our scenario, they are then trained for 20 epochs, a relative small number compared to 50 epochs in \cite{zoph2016neural}. Besides, we use a smaller initial learning rate for this reason. 
Other settings for training networks on CIFAR-10 and SVHN, are similar to \cite{huang2016densely,zoph2016neural}. Specifically, we use the SGD with a Nesterov momentum \cite{sutskever2013importance} of 0.9, a weight decay of 0.0001, a batch size of 64. The initial learning rate is 0.02 and is further annealed with a cosine learning rate decay \cite{gastaldi2017shake}. The accuracy in the held-out validation set is used to compute the reward signal for each sampled network. Since the gain of improving the accuracy from 90\% to 91\% should be much larger than from 60\% to 61\%, instead of directly using the validation accuracy $acc_v$ as the reward, as done in \cite{zoph2016neural}, we perform a non-linear transformation on $acc_v$, i.e. $\tan(acc_v \times \pi / 2)$, and use the transformed value as the reward. Additionally, we use an exponential moving average of previous rewards, with a decay of 0.95 as the baseline function to reduce the variance.

\begin{figure}[t]
    \centering
	\includegraphics[width=0.9\columnwidth]{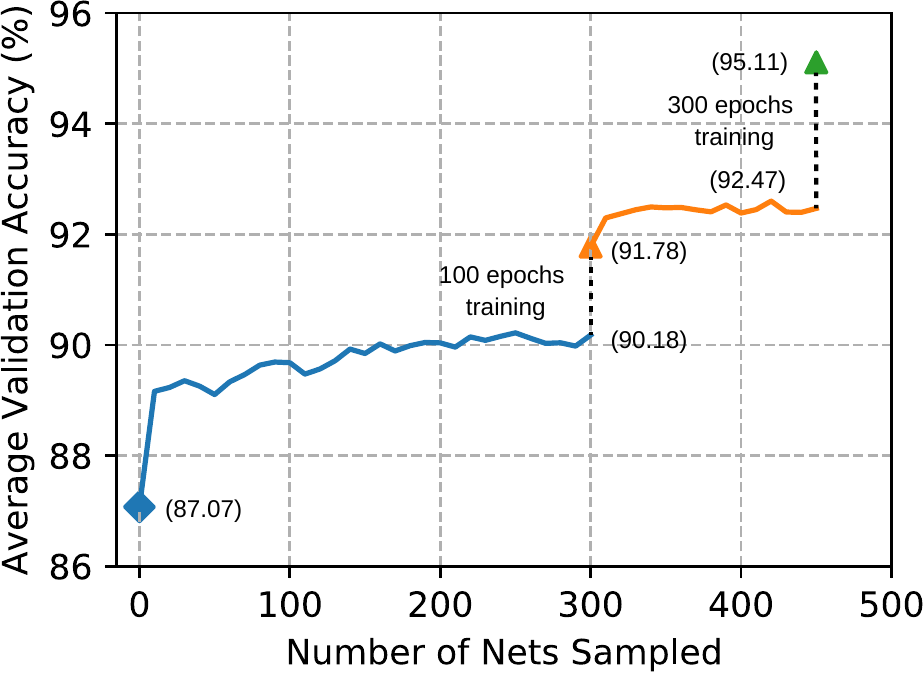}
	\caption{Progress of two stages architecture search on C10+ in the plain CNN architecture space.}
	\label{fig:cifar10_steps}
\end{figure}

\subsection{Explore Plain CNN Architecture Space}\label{para:plain_cnn_exp}
We start applying EAS to explore the plain CNN architecture space. Following the previous automatic architecture designing methods \cite{baker2016designing,zoph2016neural}, EAS searches layer parameters in a discrete and limited space. For every convolutional layer, the filter size is chosen from \{1, 3, 5\} and the number of filters is chosen from $\{16, 32, 64, 96, 128, 192, 256, 320,$ $ 384, 448, 512\}$, while the stride is fixed to be 1 \cite{baker2016designing}. For every fully-connected layer, the number of units is chosen from $\{64, 128, 256, 384, 512, 640, 768, 896, 1024\}$. Additionally, we use ReLU and batch normalization for each convolutional or fully-connected layer. For SVHN, we add a dropout layer after each convolutional layer (except the first layer) and use a dropout rate of 0.2 \cite{huang2016densely}.
%Our agent can take Net2Wider and Net2Deeper actions. For Net2Wider actions, the number of filters for convolutional layers or the number of units for fully-connected layers would be extended to the least number that is larger then the currenct number. The number of filters lay in the set $\{6, 8, 10, \ldots, 32\}$, and the number of units lay in the set $\{16, 18, 20, \ldots, 42\}$. For example, if the current number of filters is $16$, it would be extended to $18$ after taking the Net2Wider action.
% Our experiments are conducted using Tensorflow \cite{abadi2016tensorflow} with 5 GeForce GTX 1080 GPUs. 

\subsubsection{Start with Small Network} %\jun{too long one line. the same for the rest???} 
We begin the exploration on C10+, using a small network (see Table \ref{tab:start_net}), which achieves 87.07\% accuracy in the held-out validation set, as the start point. Different from \cite{zoph2016neural,baker2016designing}, EAS is not restricted to start from empty and can flexibly use any discovered architecture as the new start point. As such, to take the advantage of such flexibility and also reduce the search space for saving the computational resources and time, we divide the whole architecture search process into two stages where we allow the meta-controller to take 5 steps of Net2Deeper action and 4 steps of Net2Wider action in the first stage. After 300 networks are sampled, we take the network which performs best currently and train it with a longer period of time (100 epochs) to be used as the start point for the second stage. Similarly, in the second stage, we also allow the meta-controller to take 5 steps of Net2Deeper action and 4 steps of Net2Wider action and stop exploration after 150 networks are sampled. 

% Since our method , it is possible to divide the whole architecture search process into several stages where the best architecture from previous stages is used as the start point for the current stage, which makes each stage easier for the meta-controller to learn and search, since the agent do not have to take a very long sequence of actions to get a reward signal. And therefore  a lot resources and time.

%In this experiment, we divide the whole process into two stages. At the first stage, we allow the meta-controller to take 5 steps of Net2Deeper action and 4 steps of Net2Wider action, starting from the start net 1. After 300 networks are sampled, we take the network which achieves the best validation accuracy currently and train it with a longer period of time (100 epochs) to be used as the start point for the second stage. 

The progress of the two stages architecture search is shown in Figure~\ref{fig:cifar10_steps}, where we can find that EAS gradually learns to pick high performance architectures at each stage. As EAS takes function-preserving transformations to explore the architecture space, we can also find that the sampled architectures consistently perform better than the start point network at each stage. Thus it is usually ``safe'' to explore the architecture space with EAS. 
We take the top networks discovered during the second stage and further train the networks with 300 epochs using the full training set. Finally, the best model achieves 95.11\% test accuracy (i.e. 4.89\% test error rate). Furthermore, to justify the transferability of the discovered networks, we train the top architecture (95.11\% test accuracy) on SVHN from random initialization with 40 epochs using the full training set and achieves 98.17\% test accuracy (i.e. 1.83\% test error rate), better than both human-designed and automatically designed architectures that are in the plain CNN architecture space (see Table~\ref{tab:vs_pure}). 

We would like to emphasize that the required computational resources to achieve this result is much smaller than those required in \cite{zoph2016neural,real2017large}. Specifically, it takes less than 2 days on 5 GeForce GTX 1080 GPUs with totally 450 networks trained to achieve 4.89\% test error rate on C10+ starting from a small network. 

\begin{table}[t]
	\centering
	\caption{Test error rate (\%) comparison with CNNs that use convolutional, fully-connected and pooling layers alone.}\label{tab:vs_pure}
	\resizebox{\columnwidth}{!}{  
		\begin{tabular}{c | l | c | c}
			\hline
			& Model & C10+ & SVHN\\
			\hline
            \tabincell{l}{human \\ designed} & \tabincell{l}{Maxout \cite{goodfellow2013maxout} \\ NIN \cite{lin2013network} \\ All-CNN \cite{springenberg2014striving} \\ VGGnet \cite{simonyan2014very}} & \tabincell{c}{9.38 \\ 8.81 \\ 7.25 \\ 7.25} & \tabincell{c}{2.47 \\ 2.35 \\ - \\ -} \\        
			\hline
            \tabincell{l}{auto \\ designed} & \tabincell{l}{MetaQNN \cite{baker2016designing} (depth=7) \\ MetaQNN \cite{baker2016designing} (ensemble) \\ EAS (plain CNN, depth=16) \\ EAS (plain CNN, depth=20)} & \tabincell{c}{6.92 \\ - \\ 4.89 \\ \textbf{4.23}} & \tabincell{c}{- \\ 2.06 \\ 1.83 \\ \textbf{1.73}} \\
			\hline
		\end{tabular}
	}
\end{table}

\begin{table*}[t]
	\centering
	\caption{Test error rate (\%) comparison with state-of-the-art architectures.}\label{tab:vs_modern}
	\resizebox{0.7\textwidth}{!}{  
		\begin{tabular}{c | l | c | c | c }
			\hline
			& Model & Depth & Params & C10+\\
			\hline
            \tabincell{l}{human \\ designed} & \tabincell{l}{ResNet \cite{he2016deep} \\ ResNet (stochastic depth) \cite{huang2016densely} \\ Wide ResNet \cite{zagoruyko2016wide} \\ Wide ResNet \cite{zagoruyko2016wide} \\ ResNet (pre-activation) \cite{he2016identity} \\ DenseNet ($L=40, k=12$) \cite{huang2016densely} \\ DenseNet-BC ($L=100, k=12$) \cite{huang2016densely} \\ DenseNet-BC ($L=190, k=40$) \cite{huang2016densely} } & \tabincell{c}{110 \\ 1202 \\ 16 \\ 28 \\ 1001 \\ 40 \\ 100 \\ 190} & \tabincell{c}{1.7M \\ 10.2M \\ 11.0M \\ 36.5M \\ 10.2M \\ 1.0M \\ 0.8M \\ 25.6M} & \tabincell{c}{6.61 \\ 4.91 \\ 4.81 \\ 4.17 \\ 4.62 \\ 5.24 \\ 4.51 \\ \textbf{3.46}} \\        
			\hline
            \tabincell{l}{auto \\ designed} & \tabincell{l}{Large-Scale Evolution (250 GPUs)\cite{real2017large} \\ NAS (predicting strides, 800 GPUs) \cite{zoph2016neural} \\ NAS (max pooling, 800 GPUs) \cite{zoph2016neural} \\ NAS (post-processing, 800 GPUs) \cite{zoph2016neural} \\ EAS (plain CNN, 5 GPUs)} & \tabincell{c}{- \\ 20 \\ 39 \\ 39 \\ 20} & \tabincell{c}{5.4M \\ 2.5M \\ 7.1M \\ 37.4M \\ 23.4M} & \tabincell{c}{5.40 \\ 6.01 \\ 4.47 \\ \textbf{3.65} \\ 4.23}\\
			\hline
		\end{tabular}
	}
\end{table*}

\subsubsection{Further Explore Larger Architecture Space} To further search better architectures in the plain CNN architecture space, in the second experiment, we use the top architectures discovered in the first experiment, as the start points to explore a larger architecture space on C10+ and SVHN. This experiment on each dataset takes around 2 days on 5 GPUs. 

The summarized results of comparing with human-designed and automatically designed architectures that use a similar design scheme (plain CNN), are reported in Table \ref{tab:vs_pure}, where we can find that the top model designed by EAS on the plain CNN architecture space outperforms all similar models by a large margin. Specifically, comparing to human-designed models, the test error rate drops from 7.25\% to 4.23\% on C10+ and from 2.35\% to 1.73\% on SVHN. While comparing to MetaQNN, the Q-learning based automatic architecture designing method, EAS achieves a relative test error rate reduction of 38.9\% on C10+ and 16.0\% on SVHN. We also notice that the best model designed by MetaQNN on C10+ only has a depth of 7, though the maximum is set to be 18 in the original paper \cite{baker2016designing}. We suppose maybe they trained each designed network from scratch and used an aggressive training strategy to accelerate training, which resulted in many networks under performed, especially for deep networks. Since we reuse the weights of pre-existing networks, the deep networks are validated more accurately in EAS, and we can thus design deeper and more accurate networks than MetaQNN. 

We also report the comparison with state-of-the-art architectures that use advanced techniques such as skip-connections, branching etc., on C10+ in Table~\ref{tab:vs_modern}. Though it is not a fair comparison since we do not incorporate such advanced techniques into the search space in this experiment, we still find that the top model designed by EAS is highly competitive even comparing to these state-of-the-art modern architectures. Specifically, the 20-layers plain CNN with 23.4M parameters outperforms ResNet, its stochastic depth variant and its pre-activation variant. It also approaches the best result given by DenseNet. When comparing to automatic architecture designing methods that incorporate skip-connections into their search space, our 20-layers plain model beats most of them except NAS with post-processing, that is much deeper and has more parameters than our model. Moreover, we only use 5 GPUs and train hundreds of networks while they use 800 GPUs and train tens of thousands of networks.

\begin{figure}[t]
    \centering
	\includegraphics[width=\columnwidth]{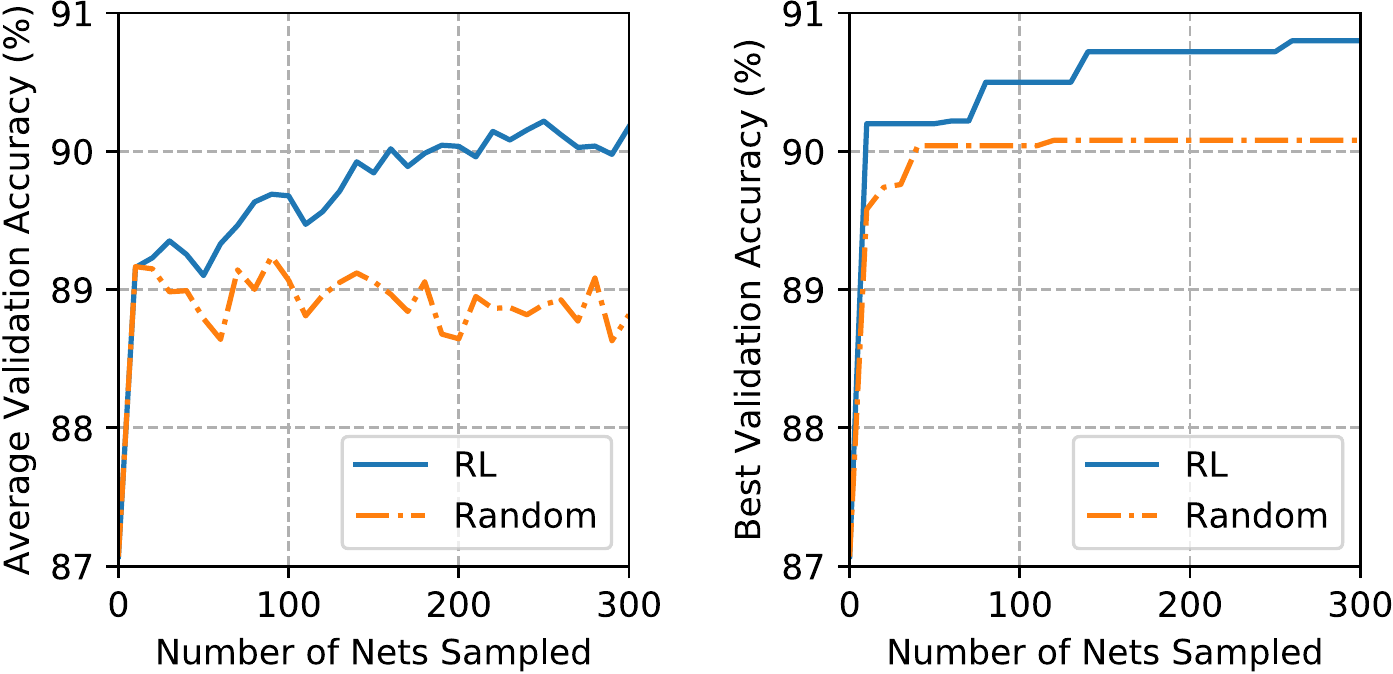}
	\caption{Comparison between RL based meta-controller and random search on C10+.}
	\label{fig:cifar10_agent_vs_random}
\end{figure}

\begin{table}[t]
	\centering
	\caption{Test error rate (\%) results of exploring DenseNet architecture space with EAS.}\label{tab:eas_with_densenet}
	\resizebox{\columnwidth}{!}{  
		\begin{tabular}{l | c | c | c | c }
			\hline
			Model & Depth & Params & C10 & C10+ \\
			\hline
            DenseNet ($L=100, k=24$) & 100 & 27.2M & 5.83 & 3.74 \\
            DenseNet-BC ($L=250, k=24$)  & 250 & 15.3M & 5.19 & 3.62 \\
            DenseNet-BC ($L=190, k=40$)  & 190 & 25.6M & - & 3.46 \\
            NAS (post-processing) & 39 & 37.4M & - & 3.65 \\
            \hline
            EAS (DenseNet on C10) & 70 & 8.6M & \textbf{4.66} & - \\
            EAS (DenseNet on C10+) & 76 & 10.7M & - & \textbf{3.44} \\
			\hline
		\end{tabular}
	}
\end{table}

\subsubsection{Comparison Between RL and Random Search} Our framework is not restricted to use the RL based meta-controller. Beside RL, one can also take network transformation actions to explore the architecture space by random search, which can be effective in some cases \cite{bergstra2012random}. In this experiment, we compare the performances of the RL based meta-controller and the random search meta-controller in the architecture space that is used in the above experiments.  
Specifically, we use the network in Table~\ref{tab:start_net} as the start point and let the meta-controller to take 5 steps of Net2Deeper action and 4 steps of Net2Wider action. 
The result is reported in Figure~\ref{fig:cifar10_agent_vs_random}, which shows that the RL based meta-controller can effectively focus on the right search direction, while the random search cannot (left plot), and thus find high performance architectures more efficiently than random search.

%Due to resource constraints, we conduct experiment versus random search in a smaller architecture space. Specifically, the number of filters is chosen from $[6, 8, \cdots, 32]$ and the number of units is chosen from $[16, 18, \cdots, 42]$. We use the start net 2 in Table \ref{tab:start_net} with validation accuracy 36.44\% as the start point and report performance comparison between policy gradient and random search in Figure \ref{fig:cifar10_agent_vs_random}. The results show that, the policy gradient can efficiently focus on the right search direction, while random search cannot (left plot), and therefore find high performance architectures much more quickly than random search. 

%\paragraph{Transfer Learning Ability:} 

%\begin{table*}[h]
%	\centering
%	\caption{Top architectures discovered during our experiments on CIFAR-10.}\label{tab:top_net}
%	\resizebox{0.8\textwidth}{!}{  
%		\begin{tabular}{| l | c | c |}
%			\hline
%			Model Architecture & Params & Test Error Rate (\%)\\
%			\hline
%			\tabincell{l}{C(192, 3, 1), C(128, 5, 1), C(96, 5, 1), C(96, 3, 1), P(2, 2), C(192, 3, 1), \\ C(256, 5, 1), C(128, 5, 1), P(2, 2), C(192, 5, 1), C(128, 3, 1), \\ C(96, 5, 1), P(2, 2), C(192, 3, 1), FC(192), SM(10)} & 5.12M & 6.74 \\
%			\hline
%			\tabincell{l}{C(192, 3, 1), C(128, 5, 1), C(256, 3, 1), C(128, 5, 1), C(128, 3, 1), P(2, 2), \\ C(256, 3, 1), C(384, 5, 1), C(256, 3, 1), C(256, 5, 1), P(2, 2), C(384, 5, 1), \\ C(384, 3, 1), C(256, 3, 1), C(256, 5, 1), P(2, 2), C(384, 3, 1), C(256, 3, 1), \\ FC(1024), FC(256), SM(10)} & 19.69M & 5.70\\
%			\hline
%		\end{tabular}
%	}
%\end{table*}

\subsection{Explore DenseNet Architecture Space}
We also apply EAS to explore the DenseNet architecture space. We use the DenseNet-BC ($L=40, k=40$) as the start point. The growth rate, i.e. the width of the non-bottleneck layer is chosen from $\{40, 44, 48, 52, 56, 60, 64\}$, and the result is reported in Table~\ref{tab:eas_with_densenet}. We find that by applying EAS to explore the DenseNet architecture space, we achieve a test error rate of 4.66\% on C10, better than the best result, i.e. 5.19\% given by the original DenseNet while having 43.79\% less parameters. On C10+, we achieve a test error rate of 3.44\%, also outperforming the best result, i.e. 3.46\% given by the original DenseNet while having 58.20\% less parameters. 

\section{Conclusion}
In this paper, we presented EAS, a new framework toward economical and efficient architecture search, where the meta-controller is implemented as a RL agent. It learns to take actions for network transformation to explore the architecture space. By starting from an existing network and reusing its weights via the class of function-preserving transformation operations, EAS is able to utilize knowledge stored in previously trained networks and take advantage of the existing successful architectures in the target task to explore the architecture space efficiently. 
Our experiments have demonstrated EAS's outstanding performance and efficiency compared with several strong baselines. For future work, we would like to explore more network transformation operations and apply EAS for different purposes such as searching networks that not only have high accuracy but also keep a balance between the size and the performance.

%In this paper, we present a RL agent, learning to take network transformation actions to explore the architecture space for automatic architecture designing.  Experiments on designing CNN architectures for an image classification task with the benchmark dataset, CIFAR-10, demonstrated the flexibility, efficiency and effectiveness of our proposed methods compared to existing automatic architecture designing approaches which train each sampled network from scratch. For future work, we would like to further explore various network transformation actions and learn RL agents for different purposes, e.g. searching networks that not only have high accuracy but also keep a balance between the size and the performance. Also, since our agent can pick any existing network as the start point, we would also explore more complex schemes for selection the start point, such as a combination of RL based approaches and neuro-evolution approaches as discussed in Section \ref{sec:connection}.  

\section{Acknowledgments}
This research was sponsored by Huawei Innovation Research Program, NSFC (61702327) and Shanghai Sailing Program (17YF1428200).

{\small
\bibliographystyle{aaai}
\bibliography{eas-trans}

\begin{thebibliography}{}

\bibitem[\protect\citeauthoryear{Bahdanau, Cho, and
  Bengio}{2014}]{bahdanau2014neural}
Bahdanau, D.; Cho, K.; and Bengio, Y.
\newblock 2014.
\newblock Neural machine translation by jointly learning to align and
  translate.
\newblock {\em ICLR}.

\bibitem[\protect\citeauthoryear{Baker \bgroup et al\mbox.\egroup
  }{2017}]{baker2016designing}
Baker, B.; Gupta, O.; Naik, N.; and Raskar, R.
\newblock 2017.
\newblock Designing neural network architectures using reinforcement learning.
\newblock {\em ICLR}.

\bibitem[\protect\citeauthoryear{Bergstra and
  Bengio}{2012}]{bergstra2012random}
Bergstra, J., and Bengio, Y.
\newblock 2012.
\newblock Random search for hyper-parameter optimization.
\newblock {\em JMLR}.

\bibitem[\protect\citeauthoryear{Cai \bgroup et al\mbox.\egroup
  }{2017}]{cai2017real}
Cai, H.; Ren, K.; Zhang, W.; Malialis, K.; Wang, J.; Yu, Y.; and Guo, D.
\newblock 2017.
\newblock Real-time bidding by reinforcement learning in display advertising.
\newblock In {\em WSDM}.

\bibitem[\protect\citeauthoryear{Chen, Goodfellow, and
  Shlens}{2015}]{chen2015net2net}
Chen, T.; Goodfellow, I.; and Shlens, J.
\newblock 2015.
\newblock Net2net: Accelerating learning via knowledge transfer.
\newblock {\em ICLR}.

\bibitem[\protect\citeauthoryear{Domhan, Springenberg, and
  Hutter}{2015}]{domhan2015speeding}
Domhan, T.; Springenberg, J.~T.; and Hutter, F.
\newblock 2015.
\newblock Speeding up automatic hyperparameter optimization of deep neural
  networks by extrapolation of learning curves.
\newblock In {\em IJCAI}.

\bibitem[\protect\citeauthoryear{Gastaldi}{2017}]{gastaldi2017shake}
Gastaldi, X.
\newblock 2017.
\newblock Shake-shake regularization.
\newblock {\em arXiv preprint arXiv:1705.07485}.

\bibitem[\protect\citeauthoryear{Goodfellow \bgroup et al\mbox.\egroup
  }{2013}]{goodfellow2013maxout}
Goodfellow, I.~J.; Warde-Farley, D.; Mirza, M.; Courville, A.; and Bengio, Y.
\newblock 2013.
\newblock Maxout networks.
\newblock {\em ICML}.

\bibitem[\protect\citeauthoryear{Han \bgroup et al\mbox.\egroup
  }{2015}]{han2015learning}
Han, S.; Pool, J.; Tran, J.; and Dally, W.
\newblock 2015.
\newblock Learning both weights and connections for efficient neural network.
\newblock In {\em NIPS}.

\bibitem[\protect\citeauthoryear{He \bgroup et al\mbox.\egroup
  }{2016a}]{he2016deep}
He, K.; Zhang, X.; Ren, S.; and Sun, J.
\newblock 2016a.
\newblock Deep residual learning for image recognition.
\newblock In {\em CVPR}.

\bibitem[\protect\citeauthoryear{He \bgroup et al\mbox.\egroup
  }{2016b}]{he2016identity}
He, K.; Zhang, X.; Ren, S.; and Sun, J.
\newblock 2016b.
\newblock Identity mappings in deep residual networks.
\newblock In {\em ECCV}.

\bibitem[\protect\citeauthoryear{Huang \bgroup et al\mbox.\egroup
  }{2017}]{huang2016densely}
Huang, G.; Liu, Z.; Weinberger, K.~Q.; and van~der Maaten, L.
\newblock 2017.
\newblock Densely connected convolutional networks.
\newblock {\em CVPR}.

\bibitem[\protect\citeauthoryear{Ioffe and Szegedy}{2015}]{ioffe2015batch}
Ioffe, S., and Szegedy, C.
\newblock 2015.
\newblock Batch normalization: Accelerating deep network training by reducing
  internal covariate shift.
\newblock {\em ICML}.

\bibitem[\protect\citeauthoryear{Kakade}{2002}]{kakade2002natural}
Kakade, S.
\newblock 2002.
\newblock A natural policy gradient.
\newblock {\em NIPS}.

\bibitem[\protect\citeauthoryear{Kingma and Ba}{2015}]{kingma2014adam}
Kingma, D., and Ba, J.
\newblock 2015.
\newblock Adam: A method for stochastic optimization.
\newblock {\em ICLR}.

\bibitem[\protect\citeauthoryear{Klein \bgroup et al\mbox.\egroup
  }{2017}]{klein2016learning}
Klein, A.; Falkner, S.; Springenberg, J.~T.; and Hutter, F.
\newblock 2017.
\newblock Learning curve prediction with bayesian neural networks.
\newblock {\em ICLR}.

\bibitem[\protect\citeauthoryear{Krizhevsky and
  Hinton}{2009}]{krizhevsky2009learning}
Krizhevsky, A., and Hinton, G.
\newblock 2009.
\newblock Learning multiple layers of features from tiny images.

\bibitem[\protect\citeauthoryear{Krizhevsky, Sutskever, and
  Hinton}{2012}]{krizhevsky2012imagenet}
Krizhevsky, A.; Sutskever, I.; and Hinton, G.~E.
\newblock 2012.
\newblock Imagenet classification with deep convolutional neural networks.
\newblock In {\em NIPS}.

\bibitem[\protect\citeauthoryear{Lin, Chen, and Yan}{2013}]{lin2013network}
Lin, M.; Chen, Q.; and Yan, S.
\newblock 2013.
\newblock Network in network.
\newblock {\em arXiv preprint arXiv:1312.4400}.

\bibitem[\protect\citeauthoryear{Mendoza \bgroup et al\mbox.\egroup
  }{2016}]{mendoza2016towards}
Mendoza, H.; Klein, A.; Feurer, M.; Springenberg, J.~T.; and Hutter, F.
\newblock 2016.
\newblock Towards automatically-tuned neural networks.
\newblock In {\em Workshop on Automatic Machine Learning}.

\bibitem[\protect\citeauthoryear{Miller, Todd, and
  Hegde}{1989}]{miller1989designing}
Miller, G.~F.; Todd, P.~M.; and Hegde, S.~U.
\newblock 1989.
\newblock Designing neural networks using genetic algorithms.
\newblock In {\em ICGA}.
\newblock Morgan Kaufmann Publishers Inc.

\bibitem[\protect\citeauthoryear{Negrinho and
  Gordon}{2017}]{negrinho2017deeparchitect}
Negrinho, R., and Gordon, G.
\newblock 2017.
\newblock Deeparchitect: Automatically designing and training deep
  architectures.
\newblock {\em arXiv preprint arXiv:1704.08792}.

\bibitem[\protect\citeauthoryear{Netzer \bgroup et al\mbox.\egroup
  }{2011}]{netzer2011reading}
Netzer, Y.; Wang, T.; Coates, A.; Bissacco, A.; Wu, B.; and Ng, A.~Y.
\newblock 2011.
\newblock Reading digits in natural images with unsupervised feature learning.
\newblock In {\em NIPS workshop on deep learning and unsupervised feature
  learning}.

\bibitem[\protect\citeauthoryear{Real \bgroup et al\mbox.\egroup
  }{2017}]{real2017large}
Real, E.; Moore, S.; Selle, A.; Saxena, S.; Suematsu, Y.~L.; Le, Q.; and
  Kurakin, A.
\newblock 2017.
\newblock Large-scale evolution of image classifiers.
\newblock {\em ICML}.

\bibitem[\protect\citeauthoryear{Schulman \bgroup et al\mbox.\egroup
  }{2015}]{schulman2015trust}
Schulman, J.; Levine, S.; Abbeel, P.; Jordan, M.~I.; and Moritz, P.
\newblock 2015.
\newblock Trust region policy optimization.
\newblock In {\em ICML}.

\bibitem[\protect\citeauthoryear{Schuster and
  Paliwal}{1997}]{schuster1997bidirectional}
Schuster, M., and Paliwal, K.~K.
\newblock 1997.
\newblock Bidirectional recurrent neural networks.
\newblock {\em IEEE Transactions on Signal Processing}.

\bibitem[\protect\citeauthoryear{Silver \bgroup et al\mbox.\egroup
  }{2016}]{silver2016mastering}
Silver, D.; Huang, A.; Maddison, C.~J.; Guez, A.; Sifre, L.; Van Den~Driessche,
  G.; Schrittwieser, J.; Antonoglou, I.; Panneershelvam, V.; Lanctot, M.;
  et~al.
\newblock 2016.
\newblock Mastering the game of go with deep neural networks and tree search.
\newblock {\em Nature}.

\bibitem[\protect\citeauthoryear{Simonyan and
  Zisserman}{2015}]{simonyan2014very}
Simonyan, K., and Zisserman, A.
\newblock 2015.
\newblock Very deep convolutional networks for large-scale image recognition.
\newblock {\em ICLR}.

\bibitem[\protect\citeauthoryear{Snoek, Larochelle, and
  Adams}{2012}]{snoek2012practical}
Snoek, J.; Larochelle, H.; and Adams, R.~P.
\newblock 2012.
\newblock Practical bayesian optimization of machine learning algorithms.
\newblock In {\em NIPS}.

\bibitem[\protect\citeauthoryear{Springenberg \bgroup et al\mbox.\egroup
  }{2014}]{springenberg2014striving}
Springenberg, J.~T.; Dosovitskiy, A.; Brox, T.; and Riedmiller, M.
\newblock 2014.
\newblock Striving for simplicity: The all convolutional net.
\newblock {\em arXiv preprint arXiv:1412.6806}.

\bibitem[\protect\citeauthoryear{Stanley and
  Miikkulainen}{2002}]{stanley2002evolving}
Stanley, K.~O., and Miikkulainen, R.
\newblock 2002.
\newblock Evolving neural networks through augmenting topologies.
\newblock {\em Evolutionary computation}.

\bibitem[\protect\citeauthoryear{Sutskever \bgroup et al\mbox.\egroup
  }{2013}]{sutskever2013importance}
Sutskever, I.; Martens, J.; Dahl, G.; and Hinton, G.
\newblock 2013.
\newblock On the importance of initialization and momentum in deep learning.
\newblock In {\em ICML}.

\bibitem[\protect\citeauthoryear{Sutskever, Vinyals, and
  Le}{2014}]{sutskever2014sequence}
Sutskever, I.; Vinyals, O.; and Le, Q.~V.
\newblock 2014.
\newblock Sequence to sequence learning with neural networks.
\newblock In {\em NIPS}.

\bibitem[\protect\citeauthoryear{Sutton and
  Barto}{1998}]{sutton1998reinforcement}
Sutton, R.~S., and Barto, A.~G.
\newblock 1998.
\newblock {\em Reinforcement learning: An introduction}.
\newblock MIT press Cambridge.

\bibitem[\protect\citeauthoryear{Williams}{1992}]{williams1992simple}
Williams, R.~J.
\newblock 1992.
\newblock Simple statistical gradient-following algorithms for connectionist
  reinforcement learning.
\newblock {\em Machine learning}.

\bibitem[\protect\citeauthoryear{Zagoruyko and
  Komodakis}{2016}]{zagoruyko2016wide}
Zagoruyko, S., and Komodakis, N.
\newblock 2016.
\newblock Wide residual networks.
\newblock {\em arXiv preprint arXiv:1605.07146}.

\bibitem[\protect\citeauthoryear{Zoph and Le}{2017}]{zoph2016neural}
Zoph, B., and Le, Q.~V.
\newblock 2017.
\newblock Neural architecture search with reinforcement learning.
\newblock {\em ICLR}.

\end{thebibliography}
}

%\appendix

\end{document}